\pdfoutput=1

\documentclass[11pt]{article}

\usepackage[]{ACL2023}

\usepackage{times}
\usepackage{latexsym}
\usepackage{booktabs}

\usepackage[T1]{fontenc}

\usepackage[utf8]{inputenc}

\usepackage{microtype}

\usepackage{inconsolata}

\usepackage{graphicx}
\usepackage{xcolor}
\usepackage{enumitem}

\newcommand{\strict}[1]{\textsc{Strict}}
\newcommand{\strictsmall}[1]{\textsc{Strict-small}}
\newcommand{\loose}[1]{\textsc{Loose}}

\title{Call for Papers - The BabyLM Challenge: Sample-efficient pretraining\\on a developmentally plausible corpus  \\ 
 \vspace{0.2cm} \url{https://babylm.github.io/}}

\author{
        Alex Warstadt \\ ETH Zürich
        \And
        Leshem Choshen \\ IBM Research 
        \And 
        Aaron Mueller \\ Johns Hopkins University
         \AND
         Ethan Wilcox \\ ETH Zürich
         \And
         Adina Williams \\ Meta AI
         \And
         Chengxu Zhuang \\ MIT
}

\begin{document}
\maketitle
\begin{abstract}

We present the call for papers for the BabyLM Challenge: Sample-efficient pretraining on a developmentally plausible corpus. This shared task is intended for participants with an interest in small scale language modeling, human language acquisition, low-resource NLP, and cognitive modeling. In partnership with CoNLL and CMCL, we provide a platform for approaches to pretraining with a limited-size corpus sourced from data inspired by the input to children. The task has three tracks, two of which restrict the training data to pre-released datasets of 10M and 100M words and are dedicated to explorations of approaches such as architectural variations, self-supervised objectives, or curriculum learning. The final track only restricts the amount of text used, allowing innovation in the choice of the data, its domain, and even its modality (i.e., data from sources other than text is welcome). We will release a shared evaluation pipeline which scores models on a variety of benchmarks and tasks, including targeted syntactic evaluations and natural language understanding. 
\end{abstract}

\section{Motivation}
\begin{figure}
    \centering
    \includegraphics[width=\linewidth]{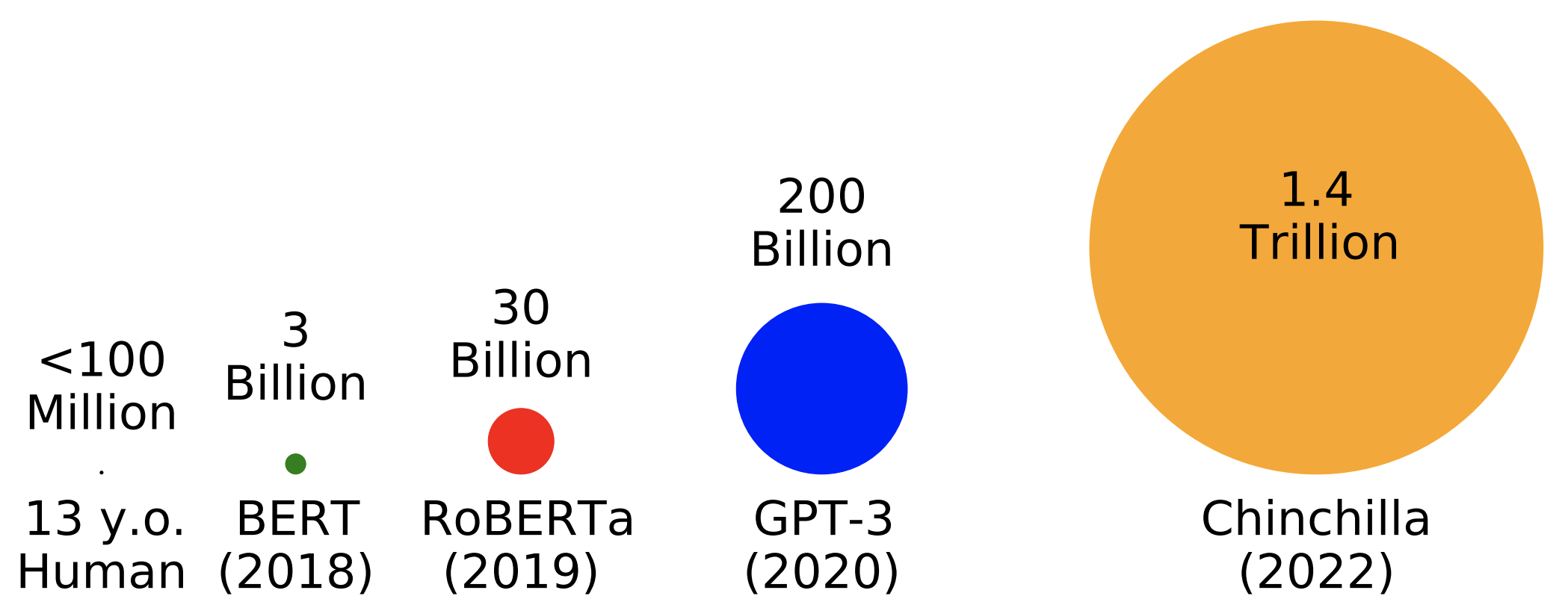}
    \caption{\textbf{Data Scale:} Modern Language Models are trained on data multiple orders of magnitude larger than the amount available to a typical human child. Image based off Fig. 1 from \citet{warstadt2022artificial}}
    \label{fig:data-scale}
\end{figure}

Huge efforts have been put into optimizing LM pretraining at massive scales in the last several years \citep{raffel-etal-2020-t5,brown-etal-2020-gpt3,chowdhery-etal-2022-palm,hoffmann2022training}. While growing parameter counts often get the most attention, datasets have also grown by orders of magnitude. 
These increasingly larger pretraining datasets are visualized, to scale, in Figure \ref{fig:data-scale}. At the same time, there has been almost no progress in pretraining at smaller human-like data scales.

Focusing on scaled-down pretraining has several potential benefits: First, small-scale pretraining can be a sandbox for developing novel techniques that improve data efficiency. These techniques have the potential to then scale up to larger datasets commonly seen in applied NLP, and could be used to enhance current approaches to modeling low-resource languages. Second, improving our ability to train LMs on the same types and quantities of data that humans learn from will give us greater access to more plausible cognitive models of humans and help us understand what allows humans to acquire language so efficiently \citep{keller-2010-cognitively,dupoux-2018-cognitive}. That is, even model failure can help in developing hypotheses about the differences between human and LM language learning.

The goal of this shared task will be to incentivize researchers with an interest in pretraining and/or cognitive modeling to focus their efforts on optimizing pretraining given data limitations inspired by human development. Additionally, we hope to democratize research on pretraining---which is typically thought to be practical only for large industry groups---by drawing attention to open problems that can be addressed on a university budget.

\section{Key Dates}

\fcolorbox{black}[HTML]{E9F0E9}{\parbox{0.48\textwidth}{
\begin{itemize}[leftmargin=0.5cm,rightmargin=0.5cm,itemsep=0em]
    \item \textbf{January 2023:} Training data released
    \item \textbf{March 2023:} Evaluation pipeline released
    \item \textbf{July 15, 2023:} Results due
    \item \textbf{August 1, 2023:} Paper submissions due
    \item \textbf{Date TBA:} Presentation at CoNLL
\end{itemize}
}}

\begin{table*}[t]
    \centering
    \resizebox{\linewidth}{!}{
    \begin{tabular}{llrrr}
    \toprule
    & & \multicolumn{2}{c}{\textbf{\# Words}} & \\\cmidrule(lr){3-4}
    Dataset & Domain & \strictsmall{} & \strict{} & Proportion \\
    \midrule
    CHILDES \citep{macwhinney2000childes} & Child-directed speech & 0.44M & 4.21M & 5\% \\
    British National Corpus (BNC),\textsuperscript{1} dialogue portion & Dialogue & 0.86M & 8.16M & 8\% \\
    Children's Book Test \citep{hill-2016-cbt} & Children's books & 0.57M & 5.55M & 6\% \\
    Children's Stories Text Corpus\textsuperscript{2} & Children's books & 0.34M & 3.22M & 3\% \\
    Standardized Project Gutenberg Corpus \citep{gerlach-2018-gutenberg} & Written English & 0.99M & 9.46M & 10\% \\
    OpenSubtitles \citep{lison-tiedemann-2016-opensubtitles2016} & Movie subtitles & 3.09M & 31.28M & 31\% \\
    QCRI Educational Domain Corpus (QED; \citealp{abdelali-etal-2014-qed}) & Educational video subtitles & 1.04M & 10.24M & 11\% \\
    Wikipedia\textsuperscript{3} & Wikipedia (English) & 0.99M & 10.08M & 10\% \\
    Simple Wikipedia\textsuperscript{4} & Wikipedia (Simple English) & 1.52M & 14.66M & 15\% \\
    Switchboard Dialog Act Corpus \citep{Stolcke-etal:2000} & Dialogue & 0.12M & 1.18M & 1\% \\
    \midrule
    \emph{Total} & -- & 9.96M & 98.04M & 100\% \\
    \bottomrule
    \end{tabular}}
    \caption{The datasets we release for the \strict{} and \strictsmall{} tracks of the BabyLM Challenge. We present the number of words in the training set of each corpus that we include. \textsuperscript{1}\url{http://www.natcorp.ox.ac.uk}\ \ \ \textsuperscript{2}\url{https://www.kaggle.com/datasets/edenbd/children-stories-text-corpus}\ \ \ \textsuperscript{3}\url{https://dumps.wikimedia.org/enwiki/20221220/}\ \ \ \textsuperscript{4}\url{https://dumps.wikimedia.org/simplewiki/20221201/}}
    \label{tab:data}
\end{table*}

\section{Tracks}

This shared task includes three tracks: \textbf{\strict{}}, \textbf{\strictsmall{}}, and \textbf{\loose{}}. 

The \strict{} and \strictsmall{} tracks require that submissions are trained exclusively on a fixed dataset, which we provide. The main difference between these tracks is the size of the dataset ($\sim$10M words vs. $\sim$100M words). Both datasets contain child-directed speech, transcribed speech from multiple sources, children's books, and Wikipedia, among other datasets. The \strictsmall{} dataset is an approximately 10\% uniform subsample of the \strict{} dataset. See \S\ref{sec:dataset} for a full description of the fixed datasets. Winners will be determined based on performance on the shared evaluation set.

The \loose{} track relaxes these restrictions. Submissions must still be trained on a maximum of 100M words, and will be tested on the shared evaluation set. However, they are permitted to use unlimited non-linguistic data or text which differs from the restricted shared task. Training on additional text is allowed without limits if that text is generated by a model trained following the above restrictions. For this track, winners will be selected holistically based on evaluation performance, relevance to the shared task goals, potential impact, and novelty.

\section{Dataset}\label{sec:dataset}

We distribute a developmentally plausible pretraining dataset inspired by the input to children.\footnote{Clicking on the following link will download the dataset (240MB zipped, 700MB unzipped): \href{https://github.com/babylm/babylm.github.io/raw/main/babylm_data.zip}{\url{https://github.com/babylm/babylm.github.io/raw/main/babylm_data.zip}}} Submissions must use only this training data to be considered for the \strict{}(\textsc{-small}) tracks, but may use different data for the \loose{} track. The dataset has two key properties:

\begin{itemize}
    \item \textbf{Under 100M words:} Children are exposed to 2M-7M words per year \citep{gilkerson2017mapping}. Choosing the beginning of adolescence (age 12) as a cutoff, the dataset should be between 24M-84M words.

    \item \textbf{Mostly transcribed speech:} Most of the input to children is spoken. Thus, we include a higher proportion of transcribed speech in our dataset.

\end{itemize}

The datasets we release are mixed domain, taken from multiple sources. Table~\ref{tab:data} summarizes the composition of the datasets.

\section{Evaluation}

We will distribute a shared evaluation pipeline based in Google Colab. Colab provides access to relatively small GPUs; this will allow users from various research settings of varying resources to efficiently evaluate their submissions. Our evaluation code will also be public, such that those wishing to use their own computational resources may do so. More details about the evaluation pipeline and the set of tasks will be released subsequently.

The pipeline assumes all models can be loaded and queried in HuggingFace’s \texttt{transformers} library \citep{wolf-etal-2020-transformers}.\footnote{While discouraged, participants whose models are not compatible with the \texttt{transformers} library can still conduct the necessary evaluation through their own pipeline.} Additionally, all models must be able to score a sequence---e.g., assign a log-likelihood or pseudo log-likelihood \cite{wang-cho-2019-bert,salazar-etal-2020-masked}---and must be able to be fine-tuned to perform classification tasks. Models do not need to be able to generate sequences. Submissions must include model outputs for each of the core evaluations in a format that we specify in our evaluation pipeline.

We choose evaluations that represent the core interests of this shared task, focusing on efficiency and applied NLP, as well as cognitive science, linguistics and language acquisition. Especially good performance in one but not both of these areas may be acknowledged with a special award.

\subsection{Baselines}
We will also release a series of baseline models with the evaluation pipeline. To train these, we simply take the hyperparameters from a series of established large language models and train them from scratch on our fixed datasets. We use hyperparameters from OPT (decoder-only; \citealp{zhang-etal-2022-opt}), RoBERTa (encoder-only; \citealp{liu-etal-2019-roberta}), and T5 (encoder-decoder; \citealp{raffel-etal-2020-t5}). These are not meant to be strong baselines, but rather to provide a naïve starting point for improving language models for this domain.

\section{Submissions}

\fcolorbox{black}[HTML]{E9F0E9}{\parbox{0.48\textwidth}{
\vspace{0.1cm}
\textbf{What you Need to Submit}
\begin{itemize}[leftmargin=0.5cm,rightmargin=0.5cm]
    \item A link where we can download the model
    \item A .zip of predictions (from our eval pipeline)
    \item A short description of the approaches taken
    \item If \loose{} track: a link where we can download any additional data
\end{itemize}
}}
\vspace{0.1cm}

Although scaled-down pretraining is more accessible to research groups with limited resources, pretraining is still expensive from a computational, energy, and financial perspective. To help groups plan for total costs, we will release an estimate of the resources required to pretrain on 10M words and 100M words. For the \loose{} track, evaluation of submissions may take into consideration computational efficiency as part of the holistic evaluation.

\section{FAQs}

\paragraph{Can papers be submitted to multiple tracks?} 
Yes. For example, a single paper can describe models which are submitted separately to the \strict{} and \strictsmall{} tracks. 

\paragraph{Can I submit a paper about my work?}
Yes, we encourage all participants to submit their reports, which will be published in the proceedings of CoNLL. You may also describe any additional experiments beyond those required for the shared task evaluation.

\paragraph{Can I submit additional evaluation metrics?}
Yes, if you wish to submit your own evaluation metrics, along with model performance, alongside our standardized evaluation results these can be considered as part of the holistic evaluation in the \loose{} track.

\paragraph{What training regimes are permitted?}
For the \strict{}/\strictsmall{} tracks, any kind of training objective/regime is permitted, as long as the data restrictions are followed. Pretrained models may not be used for any purpose such as reranking or data augmentation.

We do however require for evaluation purposes that the model provides a function to score a sequence---e.g., log-likelihood for autoregressive models or pseudo-log-likelihood for masked language models---without the need for additional fine-tuning.

\paragraph{Are there any limits on hyperparameters?}
No. In the \loose{} track, parameter efficiency and training efficiency may be considered along with other factors in ranking submissions.


\paragraph{Are there any limits on the number of epochs?}
No. We put no restrictions on the number of epochs, for several reasons: First, from an engineering perspective, training LMs with SGD tends to require multiple epochs at these scales to achieve peak performance. Second, from a cognitive perspective, humans have a memory of linguistic experience, and can continue to access and learn from these memories. Third, we try not to make a stand on implementations to allow the most freedom for innovation.

\section{Organizing Committee}

\begin{tabular}{ll}
 Leshem Choshen     & Aaron Mueller
\\ Ryan Cotterell   & Alex Warstadt
\\ Kundan Krishna   & Ethan Wilcox
\\ Tal Linzen       & Adina Williams
\\ Haokun Liu       & Chengxu Zhuang
\end{tabular} \\

\noindent \textbf{Questions?} Feel free to contact us at the following email addresses: \\
\texttt{leshem.choshen@mail.huji.ac.il} \\
\texttt{haokunl@cs.unc.edu} \\
\texttt{amueller@jhu.edu} \\
\texttt{alexwarstadt@gmail.com} \\
\texttt{ewilcox@ethz.ch} \\
\texttt{chengxuz@mit.edu}

\bibliography{custom}
\bibliographystyle{acl_natbib}

\end{document}